\documentclass[letterpaper]{article} % DO NOT CHANGE THIS
\usepackage{aaai2026}  % DO NOT CHANGE THIS
\nocopyright
\usepackage{times}  % DO NOT CHANGE THIS
\usepackage{helvet}  % DO NOT CHANGE THIS
\usepackage{courier}  % DO NOT CHANGE THIS
\usepackage[hyphens]{url}  % DO NOT CHANGE THIS
\usepackage{graphicx} % DO NOT CHANGE THIS
\usepackage{natbib}  % DO NOT CHANGE THIS AND DO NOT ADD ANY OPTIONS TO IT
\usepackage{caption} % DO NOT CHANGE THIS AND DO NOT ADD ANY OPTIONS TO IT
\usepackage{algorithm}
\usepackage{algorithmic}
\usepackage{booktabs} 
\usepackage{times}
\usepackage{subcaption}
\usepackage{newfloat}
\usepackage{listings}
\DeclareCaptionStyle{ruled}{labelfont=normalfont,labelsep=colon,strut=off} % DO NOT CHANGE THIS
\floatstyle{ruled}
\newfloat{listing}{tb}{lst}{}
\floatname{listing}{Listing}
\title{Towards Reliable Multi-Agent Systems for Marketing Applications via Reflection, Memory, and Planning}
\author {
    Lorenzo Jaime Yu Flores\textsuperscript{\rm 1, \rm 2},
    Junyi Shen\textsuperscript{\rm 1},
    Goodman Gu\textsuperscript{\rm 1},
}
\affiliations {
    \textsuperscript{\rm 1}Adobe, \textsuperscript{\rm 2}MILA - Quebec AI Institute, McGill University\\
    lorenzo.flores@mila.quebec, junyis@adobe.com, ggu@adobe.com
}

\usepackage{bibentry}
\begin{document}

\maketitle

\begin{abstract}
    Recent advances in large language models (LLMs) enabled the development of AI agents that can plan and interact with tools to complete complex tasks. However, literature on their reliability in real-world applications remains limited. In this paper, we introduce a multi-agent framework for a marketing task: audience curation. To solve this, we introduce a framework called \textbf{RAMP} that iteratively plans, calls tools, verifies the output, and generates suggestions to improve the quality of the audience generated. Additionally, we equip the model with a long-term memory store, which is a knowledge base of client-specific facts and past queries. Overall, we demonstrate the use of LLM planning and memory, which increases accuracy by 28 percentage points on a set of 88 evaluation queries. Moreover, we show the impact of iterative verification and reflection on more ambiguous queries, showing progressively better recall (roughly +20 percentage points) with more verify/reflect iterations on a smaller challenge set, and higher user satisfaction. Our results provide practical insights for deploying reliable LLM-based systems in dynamic, industry-facing environments.
\end{abstract}


\section{Introduction}

In recent years, large language models (LLMs) have been increasingly used as tools in industry applications. One popular application of LLMs is in creating AI agents — LLMs that can reason, plan, and interact with other agents or external tools (e.g. web search, calculators, phone shortcuts) to accomplish complex tasks \cite{song2025browsingapibasedwebagents, shen2025shortcutsbenchlargescalerealworldbenchmark, schick2023toolformerlanguagemodelsteach, shi2024learningusetoolscooperative, yao2023reactsynergizingreasoningacting}. While these agents excel at repetitive or moderately complex tasks, they can struggle when faced with ambiguous situations \cite{ruan2023tptulargelanguagemodelbased}, which can pose risks when deploying these systems.

To this end, various work has focused on improving the reliability and transparency of these systems to mitigate risks. These include chain-of-thought \cite{yao2023reactsynergizingreasoningacting, yao2023treethoughtsdeliberateproblem, wang2023planandsolvepromptingimprovingzeroshot}, using symbolic reasoning \cite{10.1093/nsr/nwac035, colelough2025neurosymbolicai2024systematic, bougzime2025unlockingpotentialgenerativeai, li2024formalllmintegratingformallanguage}, self-reflection \cite{Renze_2024, shinn2023reflexionlanguageagentsverbal}, or fine-tuning models directly on the domain of interest. While the aforementioned work demonstrated gains on math, code generation, common sense reasoning and QA, and game environments, literature on industry applications remains relatively scarce.

\begin{figure}
    \centering
    \includegraphics[width=\linewidth]{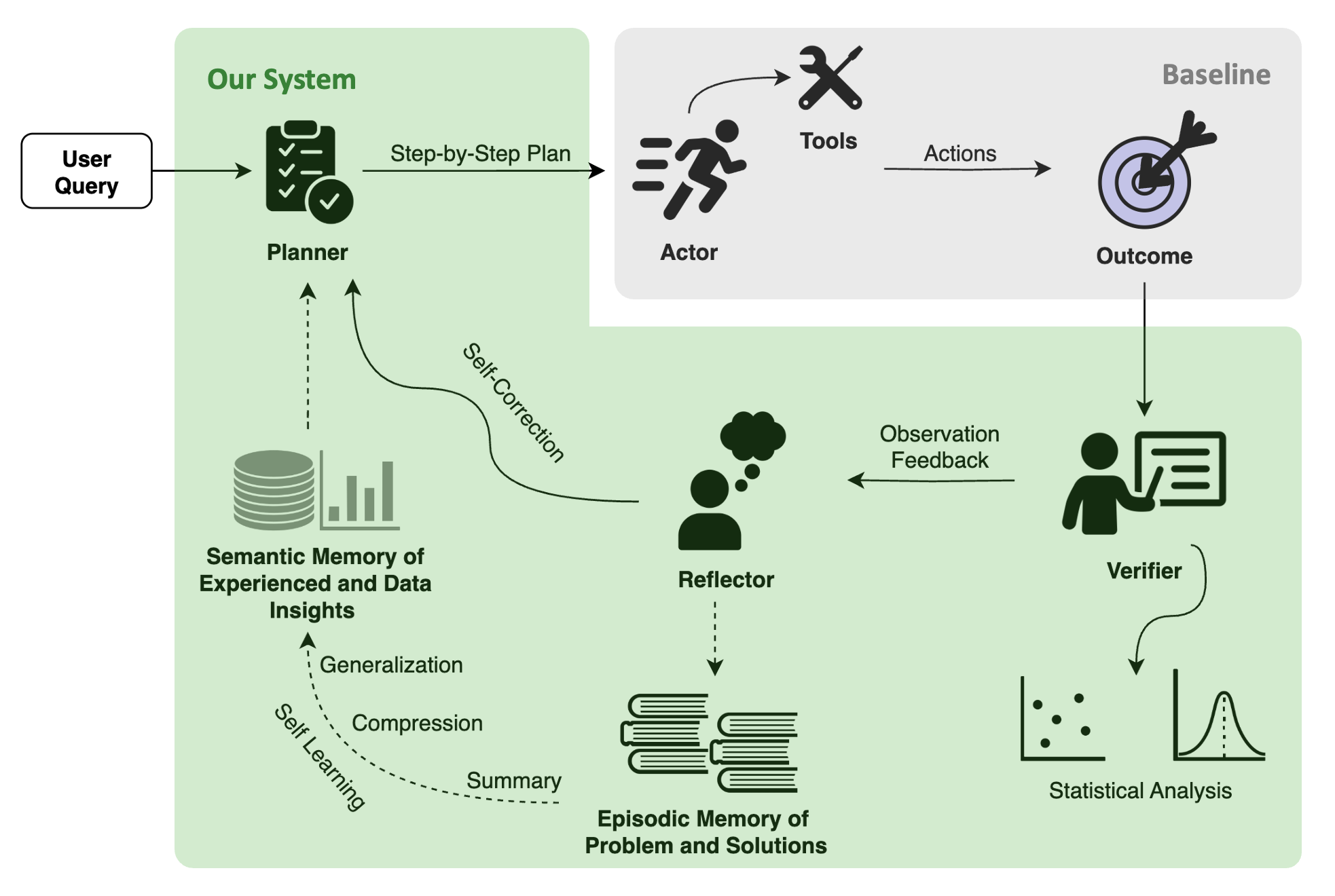}
    \caption{The audience agent consists of four modules which are responsible for modifying and planning, executing, verifying the curated audience}
    \label{Figure:overall}
\end{figure}

In this paper, we develop a framework for audience creation, where marketers and advertisers seek to select the most appropriate audience for a campaign based on user-provided natural language queries or descriptions. We develop a multi-agent system called \textbf{RAMP} (\textbf{R}eflect/Verify + \textbf{A}ct + \textbf{M}emory + \textbf{P}lan) which breaks down the task into steps assigned to specialized sub-agents (See Figure \ref{Figure:overall}). 

\begin{figure*}
    \centering
    \includegraphics[width=\linewidth]{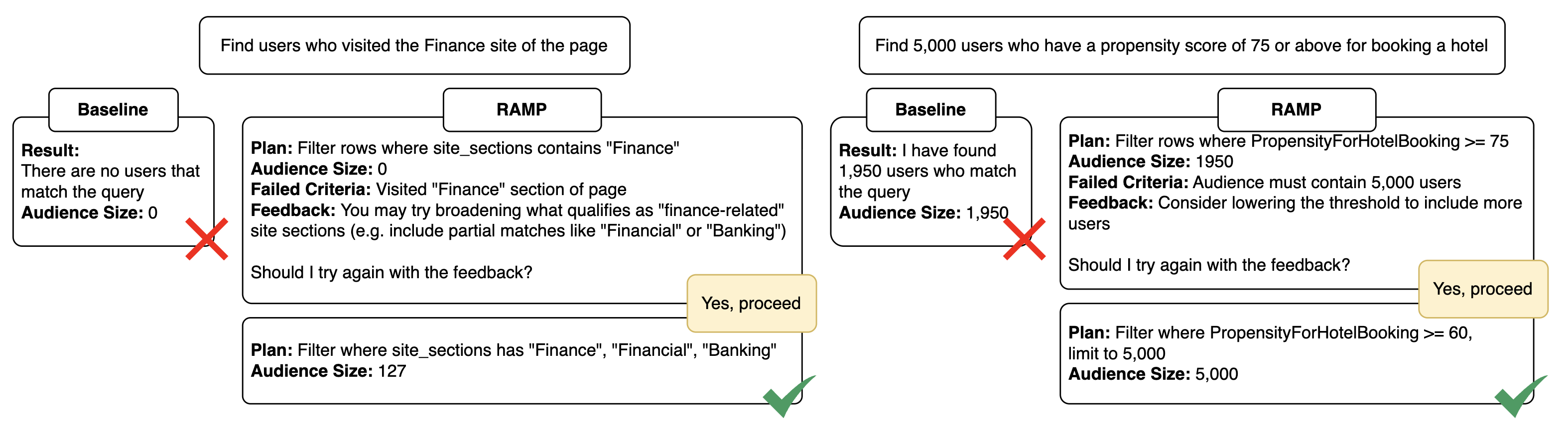}
    \caption{Our framework iteratively improves the audience created by drawing from episodic and semantic memory to propose and implement improvements to the plan}
    \label{Figure:convo}
\end{figure*}

RAMP introduces a high-level learning module that emulates human learning by compressing and generalizing insights from previous sessions. These are then used to assist a separate planning agent in refining the details of the plans that guide the actor agent in selecting the optimal tools and actions to achieve the best possible outcomes. To verify the correctness of these outcomes, we propose a neuro-symbolic verifier, which generates executable Python unit tests that check whether the selected audience meet the user criteria. When discrepancies are identified, these are passed back to the high-level learning module, which retrieves knowledge from previous sessions to determine how to address specific gaps or unmet criteria, and trigger a subsequent round of audience creation, similar to the conversation in Figure \ref{Figure:convo}.

Our investigation provides insights into the practical deployment of LLM-based agent systems. We find that providing language models with domain-specific semantic and episodic memory is essential for maintaining good performance. Without this, LLMs are unable to create better audiences, even after incorporating verification and self-reflection, and are prone to hallucinating. When episodic memory is scarce, asking the model to synthesize its own findings from the current interaction (self-learning) can help. 

We also demonstrate the value of plan generation, showing that separating the planning from the execution yields considerable improvements compared to asking the model to do both planning and execution in one LLM call. 

Finally, we find that while verification and self reflection are unnecessary for filter-based queries\footnote{More details provided in the Audience Curation Task section}, they improve performance on more ambiguous or involved queries, and boost reliablity and transparency according to users.

These findings validate and highlight the benefits of our proposed learning modules, which provide the LLMs with better context and augment its reasoning capabilities. Moreover, they demonstrate the potential of LLM agents in practical industry applications, and provide practical insights for their deployment. Our paper is organized as follows:
\begin{itemize}
    \item We first propose an audience curation task, and introduce a dataset to evaluate marketing agents;
    \item We introduce RAMP, a framework for audience curation, which incorporates verification and self-reflection modules to augment the base LLM performance
    \item We analyze the impact of each module, highlighting issues with naively applying techniques from literature, and emphasizing the need for augmenting LLMs with domain specific knowledge
\end{itemize}

\section{Related Work}

To fundamentally enhance the reasoning reliability of LLMs and AI agents, several key strategies have emerged. 

\paragraph{Planning} Previous work explore LLM's ability to generate a plan (i.e. a sequence of actions to accomplish a task), and demonstrate the benefits of using such plans to decompose complex tasks, and enhance the agent's transparency and accuracy \cite{huang2024understandingplanningllmagents, wei2025plangenllmsmodernsurveyllm}. In our work, we similarly explore the impact of plan generation before doing tool calling, in comparison to the previous baseline which does not perform explicit plan generation.

\paragraph{Neuro-Symbolic Verification}

Neuro-Symbolic (NeSy) AI use explicit intermediate representations, such as structured planning components in the model's reasoning \cite{bougzime2025unlockingpotentialgenerativeai,colelough2025neurosymbolicai2024systematic, 10.1093/nsr/nwac035}. This involves using rules, formal logic, and explicit representations of knowledge and reasoning, to perform tasks \cite{manhaeve2018deepproblogneuralprobabilisticlogic, desmet2023neuralprobabilisticlogicprogramming}. By incorporating this with neural networks, the NeSy techniques increase the reliability, interpretability, and accuracy of these networks \cite{10.1002/aaai.12149, yang2025neurosymbolicintegrationbringscausal}. Various NeSy architectures have been proposed, which employ Symbolic AI with varying degrees; based on Kautz's taxonomy \cite{Kautz_2022}, we employ the \textit{Neural} architecture, where symbolic reasoning is used by a neural network (e.g. LLMs) as a tool to process and verify the output.

In our work, we employ NL2Code verify the reliability of the output. NL2Code has been studied extensively \cite{yang2025codethinkthinkcode}, often used for solving math problems \cite{wang2023mathcoderseamlesscodeintegration, lu2024mathcoder2bettermathreasoning}, coding \cite{jain2024livecodebenchholisticcontaminationfree} and debugging \cite{jimenez2024swebenchlanguagemodelsresolve, zhang2024unifyingperspectivesnlpsoftware}, and even common sense logic and instruction-following tasks \cite{yang2025neurosymbolicintegrationbringscausal, madaan-etal-2022-language}. In our work, we use NL2Code to write Python unit tests to validate the correctness of the chosen audiences, for marketing-specific queries.

\paragraph{Self-Reflection}

Reflection is another technique for improving the accuracy and reliability of LLMs, by allowing LLMs to assess the consequences of their outputs and iteratively refine their reasoning. Through additional critiquing and reflecting of the reasoning chains, LLMs can improve the quality of their outputs \cite{lightman2023letsverifystepstep, masterman2024landscapeemergingaiagent, dutta2024autonomousagentsadaptiveplanningreasoning, Renze_2024, huang2025r2d2rememberingreflectingdynamic}. One such popular framework is Reflexion \cite{shinn2023reflexionlanguageagentsverbal}, which uses three modules: an actor, and evaluator, and a self-reflector. In this framework, the actor performs the task, such as solving a coding challenge or math problem, and the evaluator checks the correctness of the output. The self-reflector then uses both the output and the evaluator's feedback to propose improvements to the actor. In our work, extend this by splitting out the actor into planning and execution modules. This allows for more detailed feedback to be given to the planner, which helps catch errors earlier before being fed to the executor.

\paragraph{Long-Term Memory}

Various work studied how to augment LLM capabilities with \textit{memory}, which pertains to knowledge sources that store previous interactions or user-related facts \cite{jia-etal-2025-evaluating, maharana-etal-2024-evaluating, han-etal-2020-continual}. As \citet{tan-etal-2025-prospect} mention, a static memory bank may fail to adapt to new scenarios or inputs. To this end, we explore the possibility of having LLMs update their memory based on the findings from the current user session. We show the importance of client-specific memory, and the benefits of self-learning to update this memory, in controlling hallucinations and improving agent performance.

\section{Audience Curation Task}

\paragraph{Task and Benchmark} The task is as follows: given a list of 15K customers, each with 56 profile attributes, select a subset of customers (an \textit{audience}) that match a user's description (a \textit{query}). Our benchmark consists of 88 user queries and their corresponding audience from the original pool of 15K users.

To ensure objective evaluation, we restrict our user queries to filter-based queries: these queries have clear criteria that can be used to include or exclude a user from an audience. For example, ``users below 30 years old whose mailing address is in Massachusetts'', ``users who searched for Panama in the last 30 days'', or ``gold-tier loyalty users who visited the sale page'' use filter-based criteria. In contrast, goal-based queries, such as ``users who are likely to buy a car'' use more subjective criteria in selecting users.

Each of the 88 user queries correspond to 1,753 (± 2,621) users on average, out of the 15,044 candidates in the original pool. There are queries which correspond to all users, and some which correspond to none, thereby testing the robustness of a model to differing target audience lengths. The dataset also tests abilities to use various operators, such as working with dates (53/88 queries), numbers or numeric ranges (48/88 queries), and boolean values (2/88 queries).

\paragraph{Evaluation} We evaluate a model's output by computing the exact-match accuracy between the selected audiences for each of the 88 queries, and the ``gold-label'' queries, as well as average precision and recall.

\paragraph{Baselines} We compare our method to a base LLM, which only implements the planner and actor functionalities. Our aim is to analyze the impact of adding the verifier and self-reflection modules on performance.

\paragraph{Implementation} We use OpenAI GPT 4.1 with chat completions mode for all experiments. In all experiments, temperature is set to zero; all experiments are run thrice.

\section{Methodology}

We illustrate our methodology for approaching the audience selection task in Figure \ref{Figure:overall}. Our framework is patterned after Reflexion \cite{shinn2023reflexionlanguageagentsverbal}, with the executor agent split into a planner, and an actor. When a marketer inputs a query:

\begin{itemize}
    \item The planner writes detailed steps to filter the audience
    \item The actor converts the steps into executable Python functions, and runs them, to filter the raw table of customers into the desired audience
    \item The verifier then checks if the filtered audience satisfies criteria specified in the query
    \item The reflector proposes ways to modify the plan to address any criteria that were not satisfied
\end{itemize}

We describe each of the modules in more detail below, and provide the prompts and examples in the appendix.

\subsection{Planner} 

The planner is responsible for creating a detailed plan to filter the audience. This includes the columns, operations, and values for comparisons to be implemented. To aid the model, we provide it with the table metadata and a list of facts from memory. The table metadata contains column names, data types, and sample column values. The list of facts serve as \textit{semantic memory} \cite{pink2025positionepisodicmemorymissing}, which can include further descriptions of relevant columns, or notes on which operations to use based on a user query. These are retrieved from a longer list based on their similarity to the user query with BM25 \cite{10.1561/1500000019}. To implement this, we ask an LLM to return a list of steps to execute a user query, given metadata and facts (See Figure \ref{Figure:low_level_planner}). 

\subsection{Actor}

The actor executes the plan, by calling tools which filter the raw pool of customers and save the selected audience. We implement an NL2Code function, which takes in a filter in natural language, and returns a Python function called \texttt{filter\_function} which both receives and outputs a Pandas dataframe, implementing the filter. We implement the NL2Code function by calling an LLM; the Python functions are run in succession, to subset the original customer pool to the desired audience (See Figure \ref{Figure:actor} for example).

\subsection{Verifier}

The verifier is responsible for identifying audience criteria specified by the user, and verifing whether the created audience satisfy those criteria. We implement the verifier as a series of LLM calls, as shown in Figure \ref{Figure:verifier}.

First, we prompt an LLM to break down a user query into a list of individual criteria. For example, ``Give me 300 users with propensity for hotels equal or greater than 50 with lookback in the last 120 days'' can be broken down into four criteria: ``Propensity for hotels is equal to or greater than 50'', ``Propensity for hotels is less than 75'', ``The lookback period is the last 120 days'', and ``The audience has at least 300 users''. We ask the model to \textit{only} return verifiable criteria, for which there is a clear column that can be used to check the audience.

Second, we prompt a separate LLM to translate each rule into a Python function named \texttt{test\_rule}, which takes in a Pandas dataframe, and returns a boolean – which is \texttt{True} when the rule is satisfied, and \texttt{False} otherwise.

Finally, we run the functions in Python, and return a list of the rules and their execution results.

\subsection{Reflector}

The reflector proposes solutions to address the list of failed rules. For example, if the issue is that an insufficient number of customers matched the user criteria, a solution could be to relax some of the search filters applied, by expanding the geographic search area for location-based filters, or the range of values for numeric or categorical features.

To aid the model, we allow it to retrieve relevant solutions from a list of \textit{episodic memories} \cite{pink2025positionepisodicmemorymissing}, which are sentences containing an issue and a potential solution. Like the planner, we use BM25 for retrieval \cite{10.1561/1500000019}; this time, the failed rule is used as the input to BM25, to find relevant solutions (See Figure \ref{Figure:high_level_planner}).

\section{Results}

\paragraph{Adding a planning module, and providing the planner with semantic memory, significantly improve accuracy} We first focus on a one-pass attempt at the audience creation task, where we ablate equipping the model with semantic memory and adding a planner, unlike the baseline which does planning and execution in one call. Table \ref{Table:ablation1} shows that using both strategies in conjunction lead to an increase of 28 percentage points on a dataset of 88 filter-based queries.

\begin{table*}[h]\centering
    \begin{tabular}{lrrr}\toprule
\textbf{Strategy} & \textbf{Accuracy} & \textbf{Avg. Recall} & \textbf{Avg. Precision} \\
\midrule
Baseline (Actor Only) & 0.583 ± 0.004 & 0.65 ± 0.00 & 0.93 ± 0.00 \\
Actor + Planner & $^{\dagger}$0.633 ± 0.032 & $^{\ddagger}$0.71 ± 0.03 & 0.89 ± 0.01 \\
Actor + Semantic Memory & $^{\ddagger}$0.706 ± 0.023 & $^{\ddagger}$0.81 ± 0.01 & 0.94 ± 0.00 \\
Actor + Planner + Semantic Memory & $^{\ddagger}$\textbf{0.870 ± 0.021} & $^{\ddagger}$\textbf{0.95 ± 0.00} & $^{\ddagger}$\textbf{0.96 ± 0.00} \\
\bottomrule
\end{tabular}
\caption{Across 88 user queries, using a planner and adding semantic memory results in the best accuracy; Reported over 3 trials, with 2 semantic memories added; (One-sided Mann-Whitney vs Baseline, $\alpha=.10^{\dagger}, 0.05^{\ddagger}$)}
\label{Table:ablation1}
\end{table*}

Upon closer inspection, we find that models without semantic memory tend to hallucinate, despite being provided with the table metadata (e.g. column names, data type, sample values). In particular, models add unnecessary filters when querying the data, which incorrectly leads to much smaller audience sizes, as evidenced by the fewer number of IDs predicted and overall lower recall.

\begin{figure}[!h]
    \centering
    \includegraphics[width=0.9\linewidth]{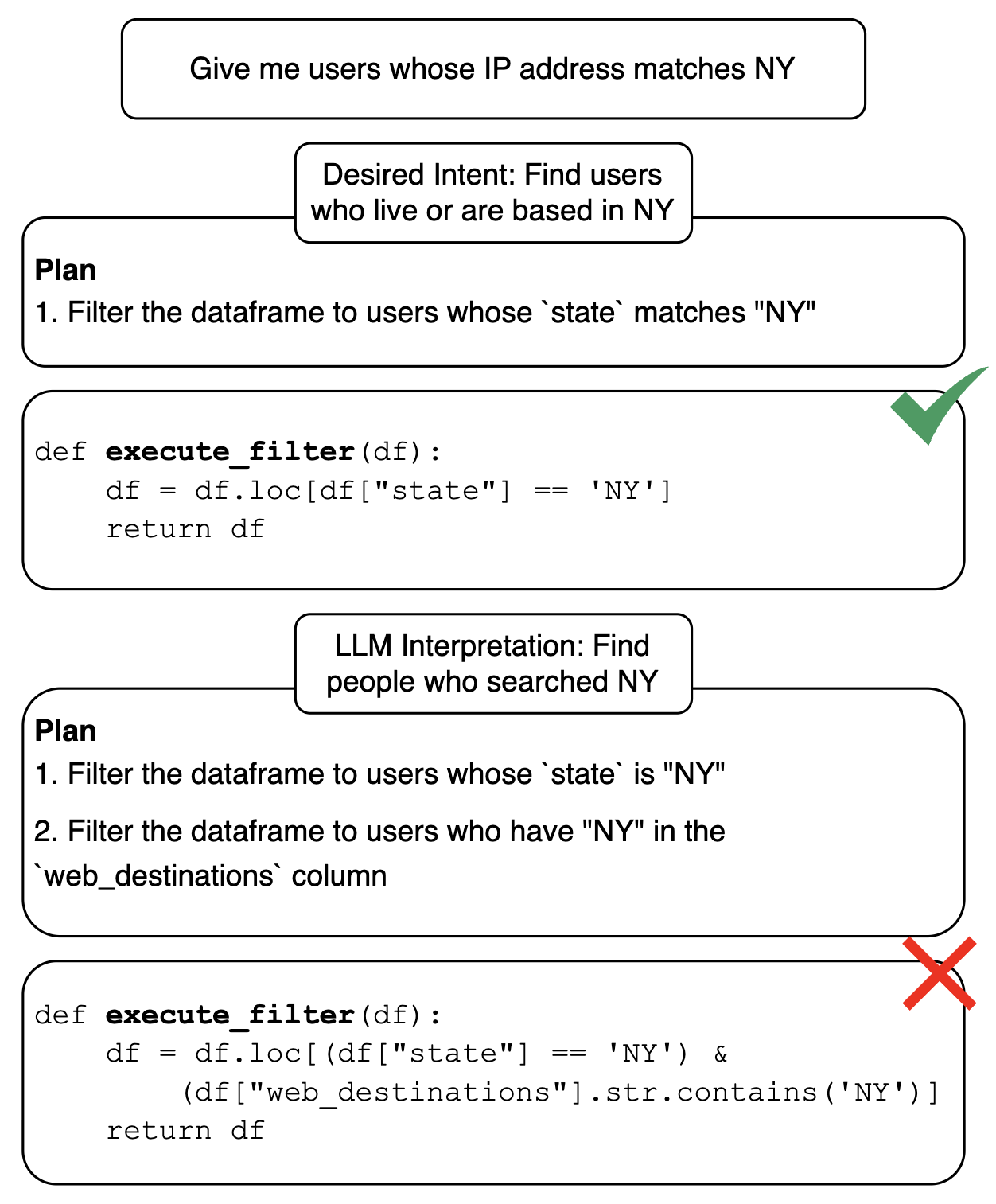}
    \caption{Unnecessary filters can lead to the wrong audience}
    \label{Figure:hallucination}
\end{figure}

For example, in Figure \ref{Figure:hallucination} the query finds users who live in NY. Filtering for users whose \texttt{state} column contains NY would suffice; however, the LLM adds a further condition to check that users also searched up NY in the \texttt{web destinations} column, leading to an incorrect audience.

\begin{figure}[htb]
    \centering
    \includegraphics[width=.95\linewidth]{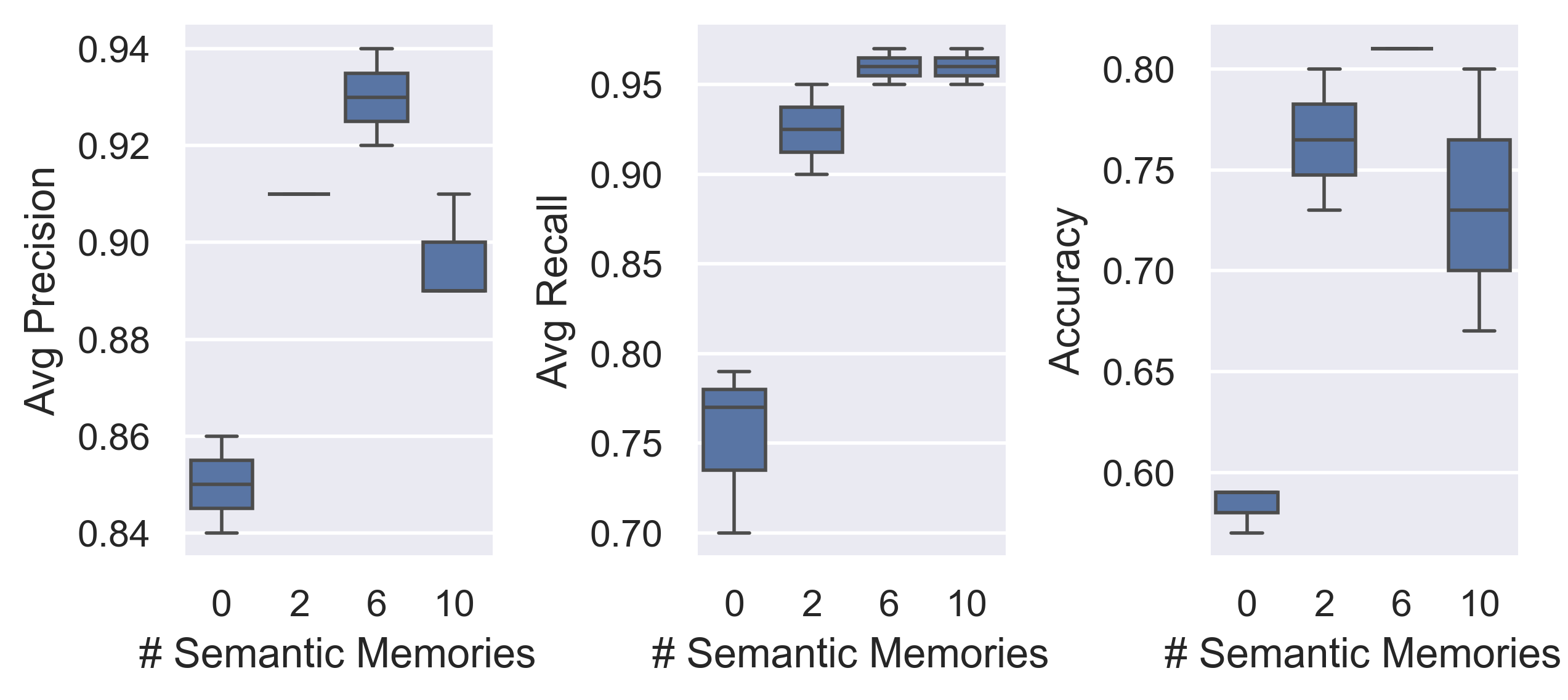}
    \caption{We find that accuracy, precision, and recall peak when 6 semantic memories are added on the 88 queries}
    \label{Figure:ablation_semantic}
\end{figure}

We study how the number of memories added impacts performance (See Fig \ref{Figure:ablation_semantic}). While recall increases with more memory, precision peaks when 6 memories are added, before dropping again. Qualitatively, we find that models hallucinate when both too little or too much memory is added. In particular, they get distracted when irrelevant memories are added, despite being prompted to ignore them, which mirrors findings from previous work that models cannot tell when context is irrelevant \cite{adlakha-etal-2024-evaluating, shi2023largelanguagemodelseasily, yoran2024makingretrievalaugmentedlanguagemodels}. This underscores the need for a good retrieval system, that only selects relevant memories.

\paragraph{On simpler filter-based queries, additional verification and self-reflection is not effective} We augment our best performing model, which uses a planner and semantic memory, with the verifier-reflector feedback loop. Here, the reflector is equipped with episodic memory. Unfortunately, we see no significant difference in the performance with and without verification (See Table \ref{Table:ablation2}). When keeping the verifier-reflector component but dropping semantic memory, we observe considerable drops in performance. Qualitatively, we observe that the model is generally able to achieve the correct audience in the first pass; additional looping then introduces other conditions or filters, that potentially degrade performance.

\begin{table*}[h]
    \centering
    \begin{tabular}{lrrrr}
        \toprule
        \textbf{Strategy} & \textbf{Accuracy} & \textbf{Avg. Recall} & \textbf{Avg. Prec.} \\
        \midrule
        Actor + Planner + Sm. Memory & \textbf{0.870 ± 0.021} & \textbf{0.95 ± 0.00} & \textbf{0.96 ± 0.00} \\
        Actor + Planner + Sm. Memory + Reflect + Verify (w/ Ep. Memory) & 0.853 ± 0.031 & 0.95 ± 0.02 & 0.96 ± 0.01 \\
        Actor + Planner + Reflect + Verify (w/o Ep. Memory) & 0.577 ± 0.045 & 0.71 ± 0.06 & 0.89 ± 0.02 \\
        Actor + Planner + Reflect + Verify (w/ Ep. Memory) & 0.713 ± 0.042 & 0.83 ± 0.06 & 0.89 ± 0.01 \\
        \bottomrule
    \end{tabular}
    \caption{Across 88 queries, the verify/reflect feedback loop does not significantly change performance; Reported over 3 trials, using 2 semantic memories added, and 1 feedback loop (One-sided Mann-Whitney vs Actor + Planner + SM $\alpha=.10^{\dagger}, 0.05^{\ddagger}$)}
    \label{Table:ablation2}
    \vspace{1em}
    \begin{tabular}{lrrr}
        \toprule
        \textbf{Strategy} & \textbf{\# Accuracy} & \textbf{Avg. Recall} & \textbf{Avg. Prec.} \\
        \midrule
        Actor + Planner + Sm. Memory & 0.133 ± 0.06 & 0.51 ± 0.05 & 0.84 ± 0.07 \\
        Actor + Planner + Sm. Memory + Reflect (no Ep. Memory) & 0.200 ± 1.83 & 0.52 ± 0.19 & \textbf{0.90 ± 0.06} \\
        Actor + Planner + Sm. Memory + Reflect (w/ Ep. Memory) & 0.233 ± 1.15 & 0.57 ± 0.22 & 0.79 ± 0.10 \\
        Actor + Planner + Sm. Memory + Verify & 0.233 ± 0.58 & $^{\dagger}$0.56 ± 0.02 & 0.84 ± 0.04 \\
        Actor + Planner + Sm. Memory + Reflect (w/ Ep. Memory) + Verify & \textbf{0.267 ± 0.58} & $^{\ddagger}$\textbf{0.64 ± 0.12} & 0.82 ± 0.02 \\
        \bottomrule
    \end{tabular}
    \caption{On 10 challenge queries, the verify/reflect feedback loop improves overall recall and accuracy; Reported over 3 trials, using 2 semantic memories, and one feedback loop (One-sided Mann-Whitney vs Actor + Planner + SM $\alpha=.10^{\dagger}, 0.05^{\ddagger}$)}
    \label{Table:ablation3}
\end{table*}

\paragraph{On the challenge queries, verification and self-reflection yield better recall, given that sufficient episodic memory is provided} We construct a set of 10, more challenging queries, wherein executing the query as is results in one of the criteria being violated. Hence, to tackle these queries, modifications to the plan are needed. 

We provide examples in Figure \ref{Figure:convo}. In the example on the left, the marketer asks for customers who have visited the Finance page; however, the page is actually listed as \textit{Financial Services}. Hence, searching only for \textit{Finance} will result in an empty audience. Hence, the keywords need to be expanded, in order to achieve the desired audience. On the right, the marketer wants 5,000 customers who have a propensity score above 75 for a hotel booking. However, there are not enough customers who meet this criteria. Hence, the model must be able to propose a fix, which is to lower the threshold, in order to meet the audience size requested.

\begin{figure*}[!h]
    \centering
    \includegraphics[width=15cm]{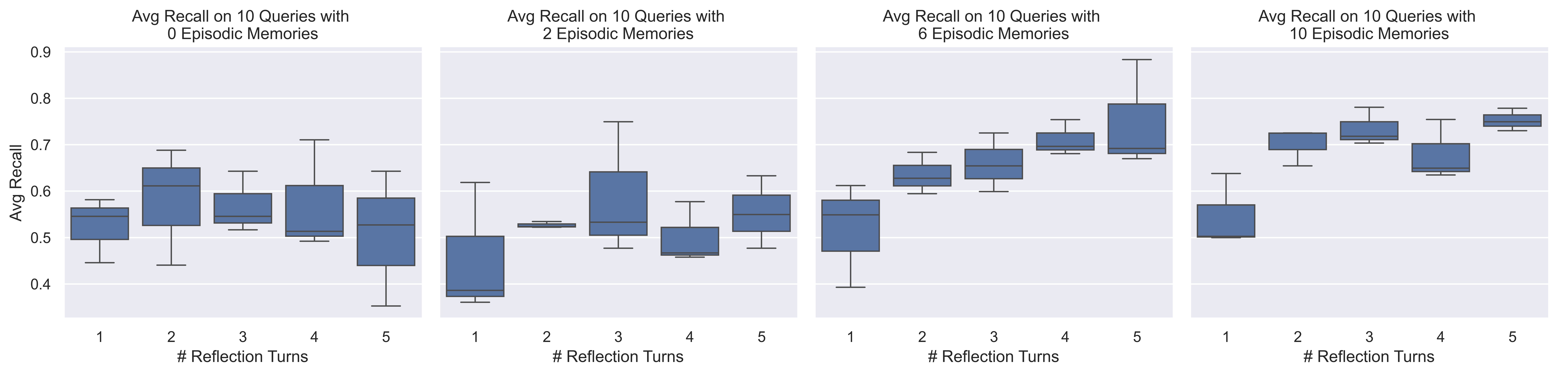}
    \caption{Recall improves with more verify/reflect steps, given ample episodic memory ($n=6,10$) (10 Challenge Queries)}
    \label{Figure:ablation_episodic_recall}
    
    \vspace{.5em}

    \includegraphics[width=15cm]{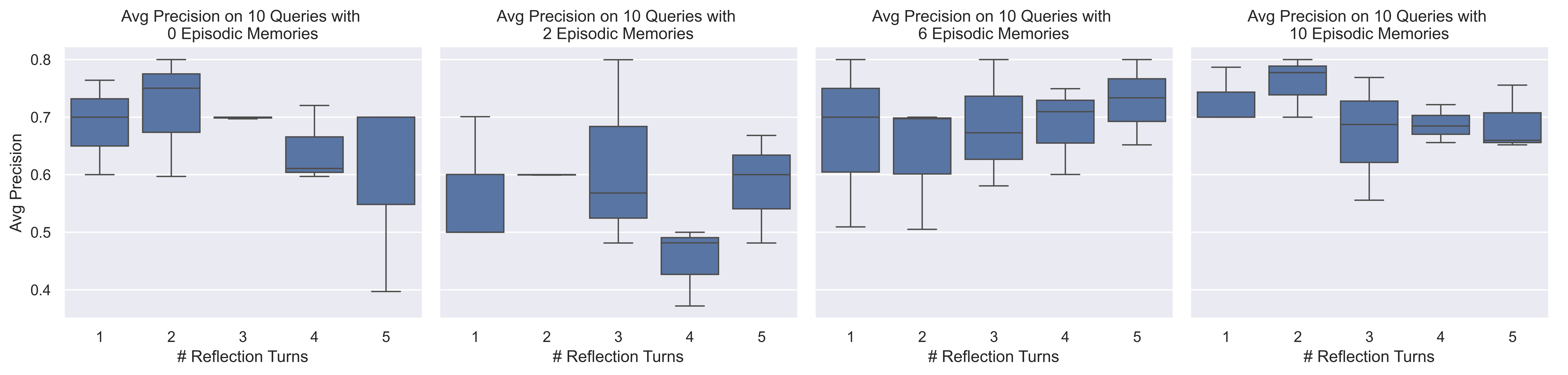}
    \caption{Based on the median, precision does not significantly deteriorate with more verify/reflect steps (10 Challenge Queries)}
    \label{Figure:ablation_episodic_precision}
\end{figure*}    

In Figure \ref{Figure:ablation_episodic_recall}, we find that more iterations generally improves the recall of the system when more episodic memory is provided. Specifically, recall for the model with 6 and 10 episodic memories progressively increase with more iterations, whereas recall for the model with only 0 or 2 episodic memories stays constant. This shows that the system can indeed modify its plans based on previously identified gaps, to capture more of the relevant audience, but only when sufficient context about previous failure cases are provided.

In addition, we find that more reflection/verification iterations does not significantly affect precision for $n=6,10$ episodic memories, at least when comparing the medians in Figure \ref{Figure:ablation_episodic_precision}. The queries we use in this round generally require some user criteria to be relaxed, in order to satisfy the user criteria. As a result, incorrect users may be added when relaxing the criteria, but fortunately this is not the case.

\begin{figure}[htb]
    \centering
    \includegraphics[width=.95\linewidth]{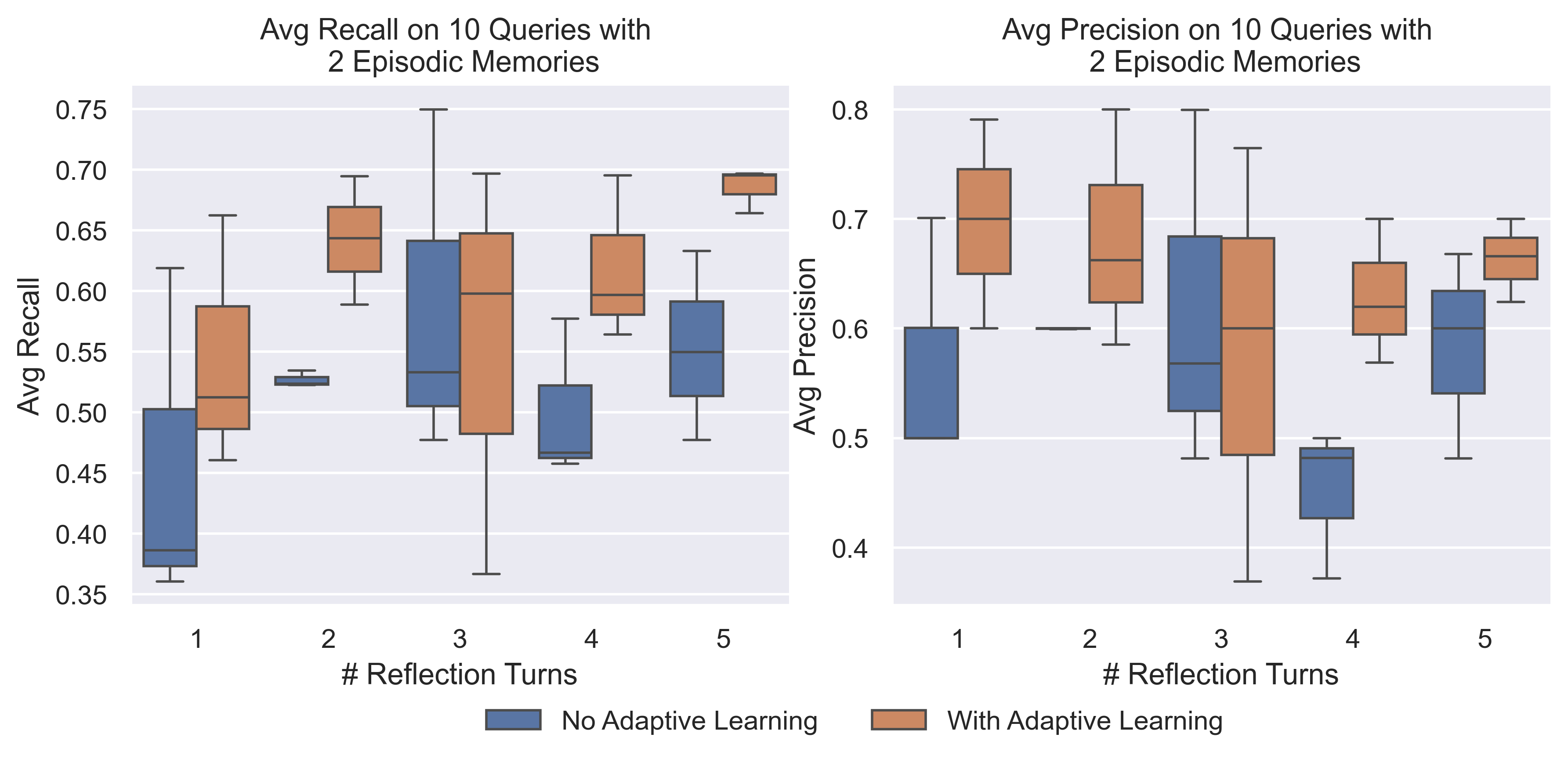}
    \caption{On challenge queries, self-learning improves recall and precision given two episodic memories are added}
    \label{Figure:ablation_episodic_with_learning_2mem}
\end{figure}

\paragraph{Allowing the model to write and use its own insights can yield better performance when there is little user-provided episodic memory} We also study the impact of letting the reflector generate its own insights to be used as semantic memory by the planner. We prompt the reflector to summarize what steps or filters should and should not be used, based on the previous plan and corresponding results. We observe that when only 2 episodic memories are provided by the user, adding LLM self-learning generally improves recall and precision (See Figure \ref{Figure:ablation_episodic_with_learning_2mem}). Hence, LLM self-learning can be useful when there is not much human written episodic memory available. However, when either no memory or too much memory has been added, self-learning can negatively impact precision and recall (See Figure \ref{Figure:ablation_episodic_with_learning}). Hence, while self-learning has the potential to improve performance, it should be applied with caution.

\paragraph{Verification and reflection yield the best performance when used together} We ablate the impact of reflection and verification on the 10 challenge queries, and find that using both together yield the biggest gain in recall (Table \ref{Table:ablation3}).

\paragraph{Users also rate the framework with reflection and verification as having higher transparency and accuracy, and better overall user experience, compared to the baseline} We survey three product managers, asking them to compare the outputs of the baseline and RAMP. We present the users with eight questions taken from the challenge set, with the other two used for practice. In each question, we ask users to choose the output they prefer with respect to three criteria. In case both are similar, they may also say ``neither''. To not unfairly bias either model, we say ``Yes, proceed with audience creation'' whenever the model elicits more information.

\begin{itemize}
    \item \textbf{Accuracy:} Which output outlines a more accurate plan, that would align better with a marketer's thinking?
    \item \textbf{Transparency:} Regardless of accuracy, which output is more clear about what it actually did with the data?
    \item \textbf{User Experience:} Which would a marketer prefer?
\end{itemize}

\begin{table}[h]\centering
    \begin{tabular}{lrrrr}\toprule
\textbf{Criterion} & \textbf{Baseline} & \textbf{Neither} & \textbf{Ours} \\
\midrule
Accuracy & 17\% & 42\% & 42\%\\
Transparency & 21\% & 0\% & 79\% \\
Overall UX & 25\% & 4\% & 71\% \\
\bottomrule
\end{tabular}
\caption{Across 8 questions, 3 PMs prefer RAMP in terms of accuracy, transparency, and overall user experience}
\label{Table:user_study}
\end{table}

As shown in Table \ref{Table:user_study}, the users prefer our framework over the baseline more frequently across all three criteria. They mentioned how the framework's ability to iterate and find audience, when the baseline could not, led to better results that were more helpful in practical situations. There were solutions that may not have occurred to them – which shows how episodic memory can augment a marketer's knowledge, based on interactions with previous marketers. Finally, they found the system more transparent, noting how the verification helped explain what criteria were used and satisfied.

One drawback of verification/reflection is that the iteration can be very tedious. Sometimes, users appreciated how to-the-point the baseline was; hence, a balance between eliciting user feedback and choosing which actions or modifications to perform autonomously needs improvement.

\section{Conclusion}

We propose a task for agent-based frameworks in the marketing domain, and demonstrate the impact of various memory and reflection based components in solving it. We find that overall, semantic and episodic memory are crucial towards system performance, whereas the reflect/verify paradigm is most useful when queries are ambiguous.

There are various directions future work can take to improve both the core framework and the user experience. We observe that there can be significant hallucination if the agent misunderstands the query (See Figure \ref{Figure:hallucination}), which both impact the accuracy and user experience, which future work can prevent. Moreover, users mention how the reflect/verify paradigm leads to long and tedious back-and-forth loops, and future work can study how to improve the efficiency of the system to make the user experience smoother.

A limitation is we only test OpenAI GPT 4.1 on a narrow domain. While the goal was to show practical considerations in using these systems, future work can explore the applicability of RAMP with other models and in other domains.

\bibliography{aaai2026}

% \section{Acknowledgments}
\clearpage

\appendix

\section{Prompts and Examples for RAMP Modules}
\label{Appendix:prompts}

\begin{figure}[htbp]
    \centering
    \includegraphics[width=\linewidth]{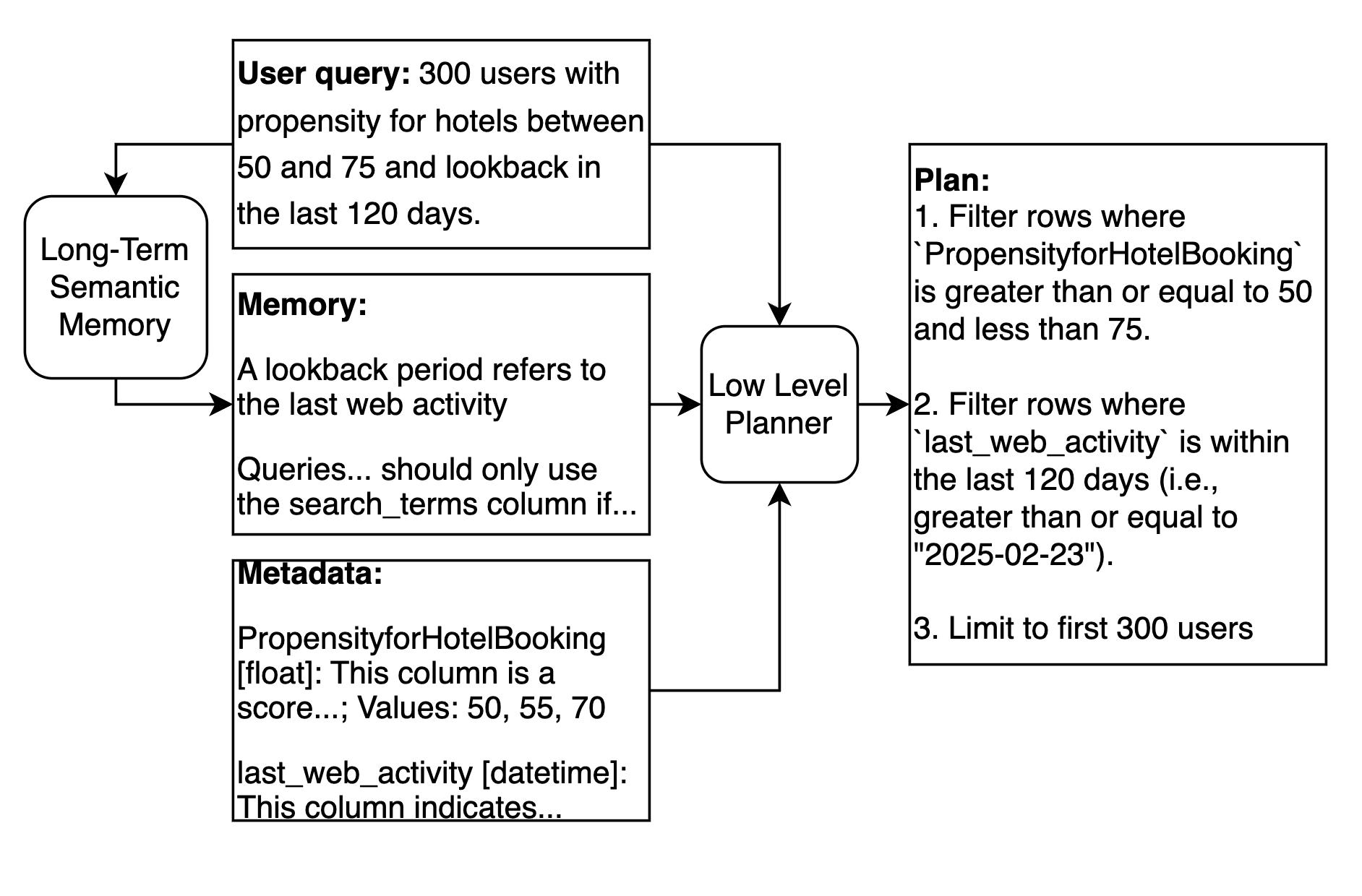}
    \caption{The planner generates a step by step plan for filtering the audience using the table metadata}
    \label{Figure:low_level_planner}
\end{figure}

\begin{figure}[htbp]
    \centering
    \includegraphics[width=\linewidth]{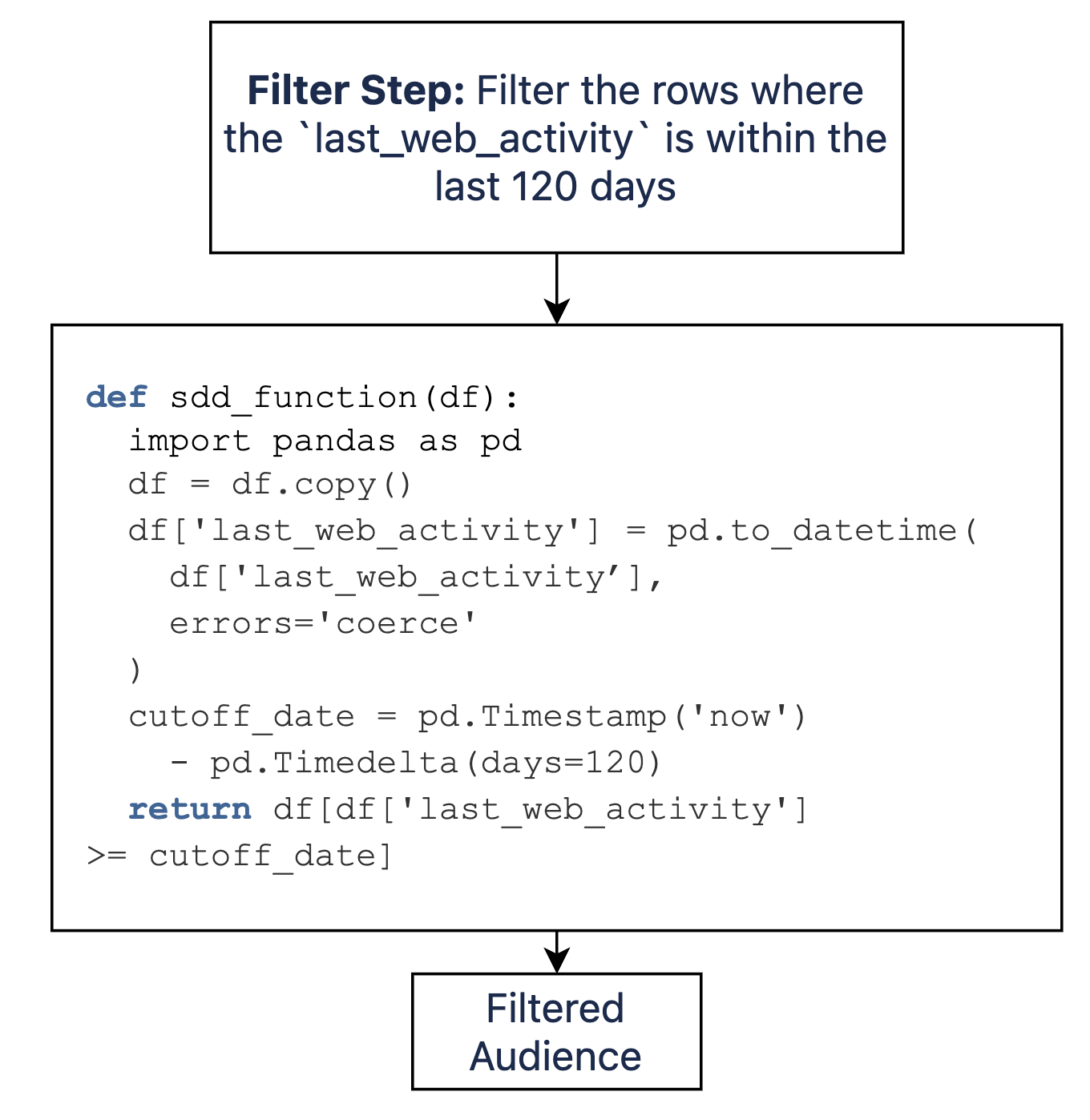}
    \caption{The actor uses an NL2Code function which translates a natural language filter into a Python function, that is then executed to generate a filtered subset of the customers}
    \label{Figure:actor}
\end{figure}

\begin{figure*}[htbp]
    \centering
    \includegraphics[width=\linewidth]{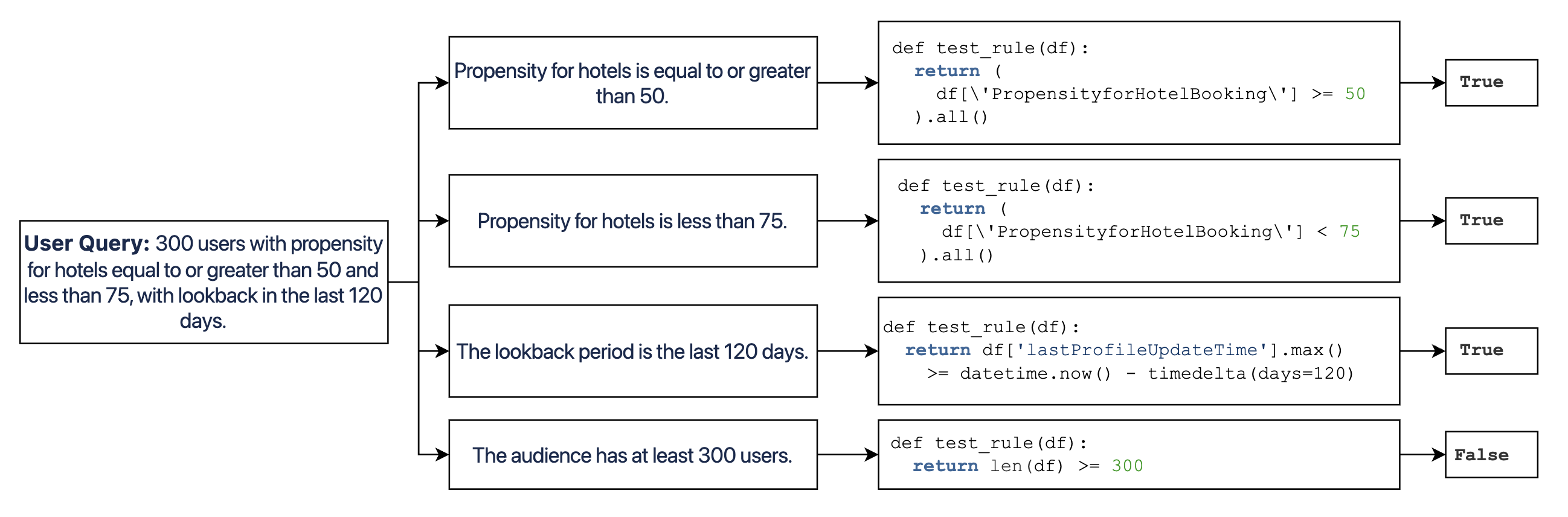}
    \caption{The verifier splits the user query into verifiable criteria, and executes Python functions to test whether they are satisfied}
    \label{Figure:verifier}
\end{figure*}

\begin{figure}[htbp]
    \centering
    \includegraphics[width=\linewidth]{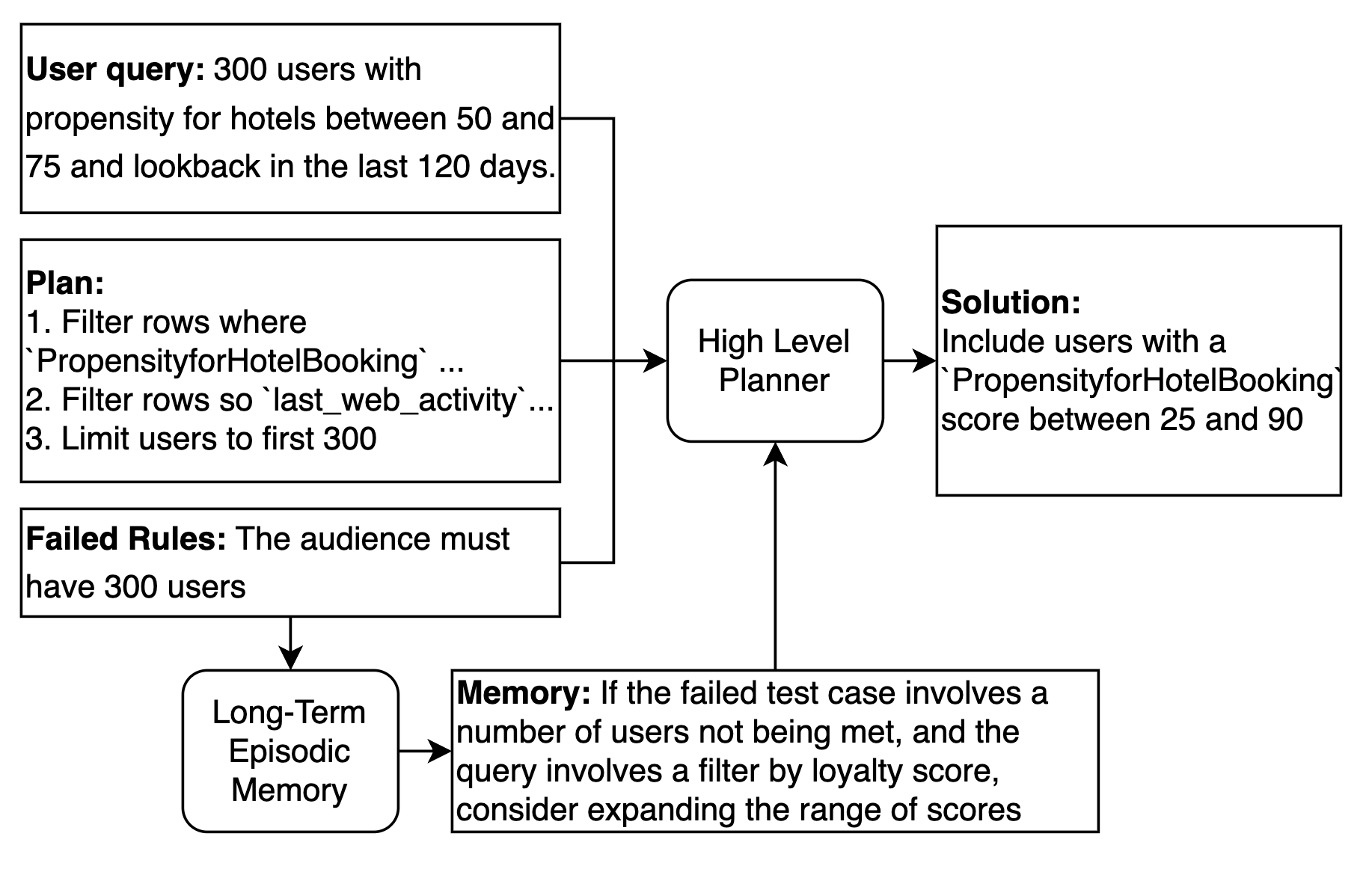}
    \caption{The reflector uses episodic memory to propose solutions to the failed user criteria}
    \label{Figure:high_level_planner}
\end{figure}

We provide the prompts for the modules in Table \ref{Table:prompts1} and \ref{Table:prompts2}.

\begin{table*}[p]
\centering
\begin{tabular}{p{3cm} p{12cm}}
\toprule
\textbf{Model} & \textbf{Example} \\
\midrule
\textbf{Planner }

(System Prompt) & You are a planner agent. Your goal is to create a plan to select the best subset from a dataset that matches the user's query. 

Use the external feedback from the failed Python unit tests, the troubleshooting suggestions from the client, and the critiquer's feedback to create an accurate plan. 

Your plan should only be a concise list of steps. 

Only include steps for applying filters or calling tools. 

Do not include steps about data cleaning, understanding the data, or post-processing.\\
\cmidrule{2-2}
\textbf{Planner}

(User Prompt) & You are a planner agent. Your goal is to create a plan to select the best subset from a dataset that matches the user's query.

Base how you will filter the data or call tools using the metadata of the dataset.
If available, use the facts from memory to guide the plan writing, but only incorporate facts that are relevant to the user's query.

User Query: \{user\_query\}

Metadata: \{metadata\}

Critiquer Feedback: \{critiquer\_feedback\}

\{memory\_prompt\}

Response Format:
[PLANNER OUTPUT]

Plan: \\
\midrule
\textbf{Verifier}

(Rule Extraction) & 

Please extract the verifiable statements from the user prompt.

The user prompt is: \{user\_prompt\}

Only include statements that are objectively verifiable, such as "The number of users is X", "The score is above X", "The date range is between X and Y".
Do not include subjective predicates, which typically include phrases like "most likely", "soonest", "latest", "high propensity", etc.
Do not include statements that are not verifiable, such as "Adhere to the quantity and location criteria", "Keep your answer concise", "Do not include any other text"
If the user prompt includes a statement like "Assume today is YYYY-MM-DD", do not include it in the verifiable statements.
Note that the number of users is also a verifiable statement.
The verifiable statements are: \\
\cmidrule{2-2}
\textbf{Verifier}

(Rule to Python Code Translation) & 

Given a Pandas dataset with the following columns and datatypes:
\{metadata\}

Relevant Facts: \{memory\}

Write a Python function named \texttt{test\_rule} that takes a Pandas dataframe \texttt{df}, and runs a test to check if \texttt{df} meets the rule.
The function should return a boolean value indicating whether the rule is met.
Think rigorously about how to test the rule, using only the minimum number of columns to verify the rule.
Do not hallucinate additional constraints or conditions.            
Return only the function definition.
The rule is: \{rule\}

The python code is:\\
\bottomrule
\end{tabular}
\caption{Prompts for Audience Segmentation Model Modules}
\label{Table:prompts1}
\end{table*}

\begin{table*}[p]
\centering
\begin{tabular}{p{2cm} p{13cm}}
\toprule
\textbf{Model} & \textbf{Example} \\
\midrule
\textbf{Reflector}

(System Prompt) & 
You are a critiquer agent. You will be presented with Review the feedback and suggest improvements to the plan.\\
\cmidrule{2-2}
\textbf{Reflector}

(User Prompt) & You are a critiquer agent for an audience targeting task. You will be presented with three things:

1. User query: describes the audience segment the user wants to create.

2. Plan: list of steps taken to filter a dataframe, to generate the desired audience.

3. List of failed test cases, and a set of facts that may or may not be relevant to the failed test cases.

Your goal is to address the failed test cases, by improving the plan. You are not supposed to suggest additional changes that do not directly contribute to fixing the failed test cases.

Remember, another agent will be executing the plan, so you should not suggest changes that are not executable by the tools.

If the facts or memory is not relevant to the failed test cases, disregard them.
Then, you will update the user query following the guidelines below.

You will return three things:

(1) a list of suggested changes to the plan,

(2) an updated user query if changes are necessary, otherwise the original user query, and

(3) a list of distilled insights from this interaction, that can be used to improve the next reflection

For the suggested changes to the plan:

- Only suggest changes that involve ways to improve how we filter the data. The tools are unable to do other operations, such as data cleaning, ensuring data quality, or some further analysis that is not related to the data filtering.

- If the failed test cases are provided, try to identify what about the plan caused the test case to fail, and use that to suggest changes to the plan.

- The potential solutions are not strict rules, use your judgement to determine whether to use them or not.

- Keep your suggestions succinct, and group changes related to the same filter into a single suggestion.

- Start each suggestion with "Consider" or "You may try", the suggestions should sound like recommendations, not commands.

For the updated user query:

- Do not add new filters or suggestions to the query; you can only drop filters

- If you suggested to change a filter, drop the filter from the user query.

    - If you suggested to change a numeric threshold, drop the numeric filter from the user query.
    
    - If you suggested to change a geographic area, drop the geographic filter from the user query.
    
    - If you suggested to change a date range, drop the date range filter from the user query.

For the distilled insights:

- Synthesize the insights from the failed test cases and possible solutions

- Add the changes that resulted in failed test cases, which the future agent can avoid repeating

- Add the changes that resulted in improved performance, which the future agent can use to improve the plan

- Mimic human learning in the way that you distill, summarize, and generalize from the failed test cases and possible solutions

User query: \{user\_query\}

Plan: \{plan\}

Feedback: \{feedback\}

Only return the suggested changes to the plan, the updated user query, and the distilled insights. \\
\bottomrule
\end{tabular}
\caption{Prompts for Audience Segmentation Model Modules}
\label{Table:prompts2}
\end{table*}

\section{Performance of Self-Learning Module}
\begin{figure*}[htb]
    \centering
    \includegraphics[width=16cm]{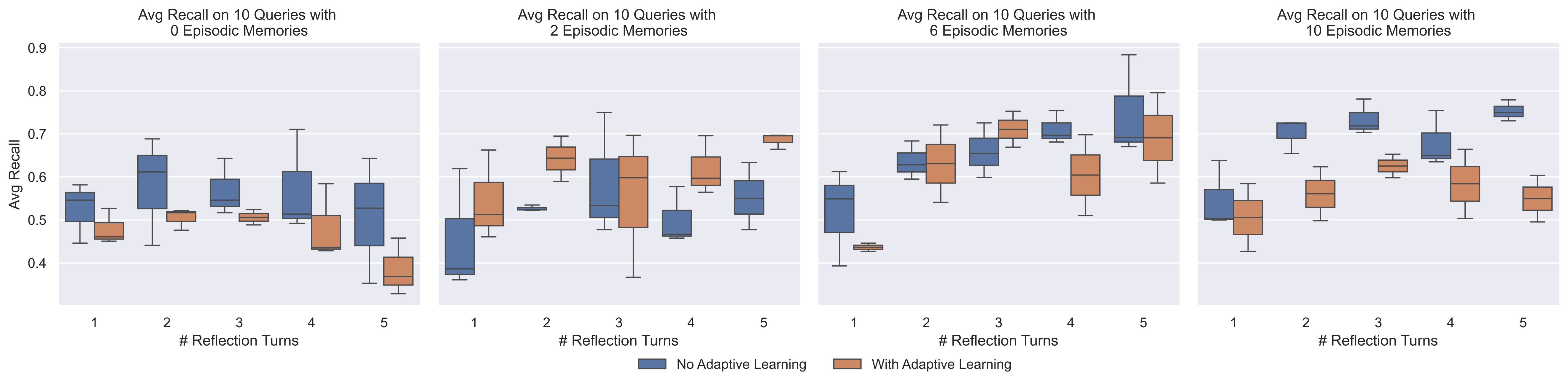}
    \caption{On the challenge queries, there are diminishing returns to adding more episodic memories to the reflector when the model is allowed to augment its own memory with its own synthesis. Overall, the verify/reflect paradigm yields directionally positive results as more iterations are done}
    
    \vspace{1em} % optional spacing between the figures

    \includegraphics[width=16cm]{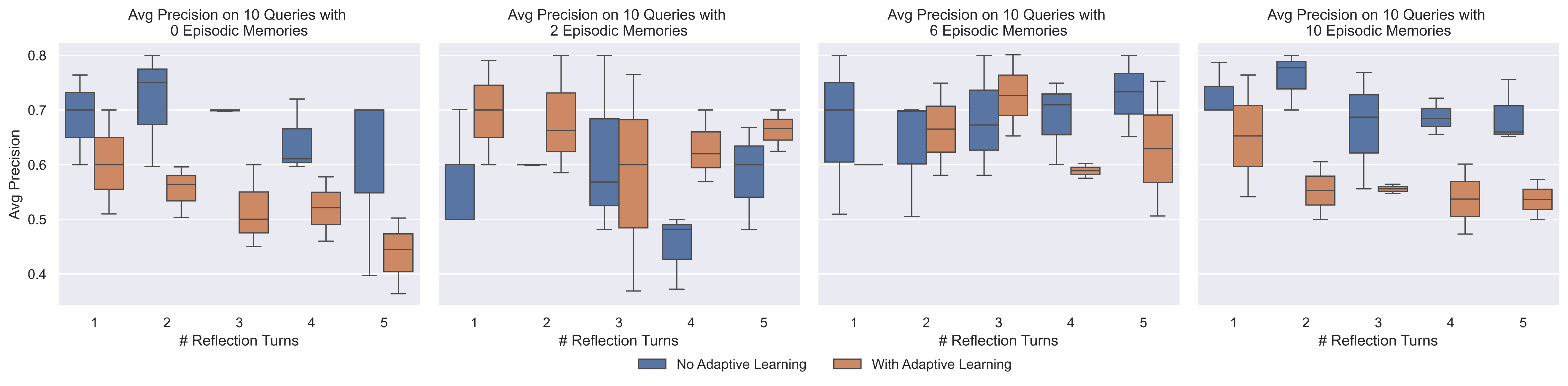}
    \caption{On the challenge queries, the precision slightly degrades with more learning when the model is allowed to write its own episodic memory.}
    \label{Figure:ablation_episodic_with_learning}
\end{figure*}

We plot the precision and recall when LLM self-learning is used in Figure \ref{Figure:ablation_episodic_with_learning}

\label{Appendix:self_learning}

\end{document}